\documentclass[runningheads]{llncs}
\usepackage[T1]{fontenc}

\usepackage{graphicx}
\usepackage{amsmath}
\usepackage{amssymb}
\usepackage{cleveref}
\usepackage{algorithm}
\usepackage{comment}
\usepackage{float}
\usepackage{booktabs}   
\usepackage{array}      
\usepackage[noend]{algpseudocode}
\usepackage{makecell}
\usepackage{tabularx}
\usepackage[table]{xcolor}

\begin{document}
\title{GymPN: A Library for Decision-Making in Process Management Systems}
%
%
\author{Riccardo Lo Bianco\inst{1, 2} \and
Remco Dijkman\inst{1, 2}\and
Willem van Jaarsveld\inst{1,2}\
}

\authorrunning{Riccardo Lo Bianco et al.}   
%
\tocauthor{Riccardo Lo Bianco}
\institute{$^1$ Eindhoven University of Technology, Netherlands\newline
$^2$ Eindhoven Artificial Intelligence Systems Institute, Netherlands\newline
\email{\{r.lo.bianco|r.m.dijkman|w.l.v.jaarsveld\}@tue.nl}
}
\maketitle              
\begin{abstract}
Process management systems support key decisions about the way work is allocated in organizations. This includes decisions on which task to perform next, when to execute the task, and who to assign the task to. Suitable software tools are required to support these decisions in a way that is optimal for the organization.
This paper presents a software library, called GymPN, that supports optimal decision-making in business processes using Deep Reinforcement Learning. GymPN builds on previous work that supports task assignment in business processes, introducing two key novelties: support for partial process observability and the ability to model multiple decisions in a business process. These novel elements address fundamental limitations of previous work and thus enable the representation of more realistic process decisions. 
We evaluate the library on eight typical business process decision-making problem patterns, showing that GymPN allows for easy modeling of the desired problems, as well as learning optimal decision policies.

\end{abstract}
\section{Introduction}

Process management systems play a crucial role in supporting organizational decision-making, particularly in determining which task to perform next, when to perform it, and who should perform it. Traditionally, such decisions have been guided by simple heuristics~\cite{kumar_dynamic_2002,amaran_simulation_2016}. However, these heuristics often fail to capture the complexity and variability of real-world business processes.

In recent years, Deep Reinforcement Learning (DRL) has emerged as an alternative to heuristics and is a promising tool to support organizational decision-making. In particular, the use of DRL to optimally assign resources to activities has been an active area of investigation~\cite{shyalika_reinforcement_2020}. Yet, DRL solutions are typically tailored to specific problem formulations, making them difficult to generalize or adapt to new decision-making problems.
To address this challenge, the Action-Evolution Petri Net (A-E PN) framework was introduced as an approach to modeling organizational decision-making problems and solving them with DRL~\cite{di_francescomarino_action-evolution_2023}. The original A-E PN framework was later extended with Attributed A-E PN~\cite{marrella_universal_2024}, introducing the use of Graph Neural Networks (GNNs) to learn policies in more complex settings. Despite these advancements, two key assumptions remained. First, the framework assumes that all information about the state of the process can be used to make a decision. This assumption is often unrealistic, as in many cases only a fraction of the process variables are visible to the decision-making algorithm. Second, the framework enables modeling only one type of decision in a business process at a time. However, this limits the expressive power of the framework, because often multiple decisions need to be made in a single business process.
To overcome these limitations, this paper presents GymPN, a Python-based software library that implements and extends the A-E PN framework to overcome these limitations. GymPN introduces two main novelties: (1) support for modeling partial observability of process state, and (2) support for modeling multiple decision types within a single business process. These novel elements allow for a realistic and flexible modeling of business process management problems while retaining the ability to learn optimal decision policies through DRL. Also, since it is released as a Python library, it is easy to install and use.
GymPN is built on top of the SimPN library~\cite{dijkman_simpn_nodate}, extending its core functionalities to support the new A-E PN elements, including the distinction between action and evolution transitions, a customizable reward structure, and utilities for training and evaluating DRL algorithms as well as heuristics. The library encapsulates the entire DRL pipeline, allowing users to define new problems and train agents with minimal configuration, thereby facilitating the discovery and analysis of optimal decision policies across a wide range of business process scenarios.
The library is evaluated by applying it to eight typical organizational decision-making problems, showing that it can learn how to solve these problems in an optimal way.

Against this background, the remainder of this work is structured as follows: \cref{sec:literature} discusses related DRL frameworks and libraries for organizational decision-making; \cref{sec:preliminaries} presents the main theoretical background of the A-E PN framework; \cref{sec:method} presents the main novelties that are introduced in GymPN based on the A-E PN framework; \cref{sec:evaluation} provides a set of typical organizational decision-making problems, showing how GymPN is capable of learning optimal assignment policies in every problem instance; \cref{sec:implementation} provides a basic overview of GymPN's main features from a software perspective; \cref{sec:evaluation} presents eight typical business process decision-making problems and shows that GymPN allows the learning of optimal solutions to these problems; \cref{sec:conclusion} presents the conclusions, reflecting on the current limitations and future research steps.

\section{Related work}\label{sec:literature}

The main objective of GymPN is to provide users with an easy way to model and solve business process decision-making problems using DRL. Inspired by the variety of successful DRL-based approaches~\cite{shyalika_reinforcement_2020,bianco_automated_2025,middelhuis_rollout-based_2025}, GymPN facilitates learning and testing business process decision-making policies, using the Proximal Policy Optimization (PPO) algorithm with minimal technical overhead. In particular, the library draws from the theory of A-E Petri nets proposed in~\cite{di_francescomarino_action-evolution_2023} and extended in~\cite{marrella_universal_2024}. It further extends this theory with novel elements that allow for the representation of partially observable environments and multiple decisions per business process.
By combining a Petri net simulation environment with DRL decision-making capabilities, this work is related to~\cite{lassoued_introducing_2024}, which adopts a similar technique with the specific objective of solving job shop scheduling problems. By contrast, GymPN focuses on business process decision-making problems. To facilitate modeling and solving business process decision-making problems, GymPN integrates simulation capabilities with the decision-making capabilities of general sequential decision-making libraries. In~\cref{table:comparison} we report some of the most popular packages for sequential decision making that were produced in an academic context, relating them to GymPN.

\begin{table}[ht]
\centering
\small
\resizebox{\textwidth}{!}{%
\begin{tabular}{>{\centering\arraybackslash}m{2.8cm} >{\centering\arraybackslash}m{3.4cm} >{\centering\arraybackslash}m{2cm} >{\centering\arraybackslash}m{2.3cm} >{\centering\arraybackslash}m{2cm} >{\centering\arraybackslash}m{2cm} >{\centering\arraybackslash}m{2.3cm}}
\toprule
\textbf{Framework} & \textbf{Algorithms} & \textbf{Ease of Use} & \textbf{Ease of Modelling} & \textbf{Agent Type} & \textbf{Multi-Action} & \textbf{Partial Obs.} \\
\midrule
Gymnasium~\cite{towers_gymnasium_2024} & DRL & High & Low & Both & Yes & Yes \\
or-gym~\cite{hubbs_or-gym_2020} & DRL & High & Low & Single & No & Yes \\
pyknow~\cite{ghlala_enhancing_2022} & Heuristics & Medium & Medium & Single & No & No \\
POMDPs.jl~\cite{egorov_pomdpsjl_2017} & Exact, Heuristics, DRL & Medium & Medium & Both & No & Yes \\
pyomo~\cite{bynum_pyomooptimization_2021} & Exact, Heuristics & Medium & Medium & Both & No & No \\
DynaPlex~\cite{akkerman_dynaplex_2022} & DRL, Heuristics & High & Low & Single & No & Yes \\
GymPN & DRL, Heuristics & High & High & Single & Yes & Yes \\
\bottomrule
\end{tabular}%
}
\caption{Comparison of Business Process Optimization Frameworks.}
\label{table:comparison}
\end{table}
Compared to the basic Gymnasium (which is used internally in GymPN), our library provides a clear set of primitives that greatly simplify the modeling of business process decision-making problems. These primitives include A-E Petri net modeling capabilities, but also popular higher-level modeling languages, such as the Business Process Model and Notation (BPMN)~\cite{rosing_complete_2014}. Moreover, in GymPN there is no need to specify the state space nor the action space, which greatly reduces the modeling effort. Compared to OR-Gym, which is essentially a collection of Gymnasium environments for operations research, GymPN offers higher flexibility in the modeling phase, while still allowing for easy training and testing of DRL policies. POMDP.jl is suitable for expressing partially observable MDPs but its primitives are only adequate for simple action types. DynaPlex, on the other hand, provides simple training and testing functions for DRL, as well as high flexibility in the types of processes that can be modeled, but modeling new environments is challenging due to the low-level coding primitives.
Other packages like pyknow and pyomo allow modeling only fully observable processes, and they do not support DRL policies. 

\section{Preliminaries}\label{sec:preliminaries}

As GymPN supports modeling and simulation of business process decision-making problems based on A-E Petri net theory, this section describes the main elements of the (attributed) A-E PN framework, including the mechanism used to extract observations from the environment for a DRL agent. For brevity, we focus on the elements that are necessary to formalize the novel elements introduced in GymPN. For a complete discussion on A-E PN, the interested reader is referred to the original publications~\cite{di_francescomarino_action-evolution_2023,marrella_universal_2024}.

The (attributed) Action-Evolution Petri net is defined as follows.

\begin{definition}[Attributed Action-Evolution Petri Net] Let \(\mathcal{T} = \{'A', 'E'\}\) be a finite \textit{set of tags} representing actions and evolutions, and \(S: \mathcal{T} \to \mathcal{T}\) a \textit{network tag} update function. An Action-Evolution Petri Net (A-E PN) is as a tuple $AEPN =(A, \mathcal{E}, P, T, F, C, G, E, I, L, l_0, \mathcal{R}, \rho_0), \mathcal{A}$, where:

\begin{itemize}
\item \(A\) is a finite set of types called \textit{attributes}.
\item \(\mathcal{E}\) is a finite set of types called \textit{color sets}.
\item \(P\) is a finite set of \textit{places}.
\item \(T\) is a finite set of \textit{transitions}.
\item \(F \subseteq P \times T \cup T \times P\) is a finite set of \textit{arcs}.
\item \(C : P \to \mathcal{E}\) is a \textit{color function} that maps each place \(p\) into a set of possible token colors. \item \(G\) is a \textit{guard function} that expresses, for each transition, the condition necessary to enable that transition.
\item \(E\) is an arc expression function that expresses, for each arc, what type of tokens should be flow in the arc.
\item \(I\) is an \textit{initialization function} which determines the network's \textit{initial marking}.
\item \(L : T \to \mathcal{T}\) is a \textit{transition tag function} that maps each transition \(t\) to a single tag. Only transitions with the same tag as the network can fire at any point in time.
\item  \(l_0 \in \mathcal{T}\) is a singleton containing the \textit{network's initial tag}, usually equal to \(E\).
\item \(\mathcal{R} : T \to (f: \mathbb{R})\) associates every transition with a reward function.
\item \(\rho_0 \in \mathbb{R}\) is the initial \textit{network reward}, usually equal to \(0\).
\item \(\mathcal{A}\) is an \textit{attribute function} that maps each place \(p\) into a set of attributes. Each token on \(p\) must have a color that is composed of the token's time and a value for each of the attributes in $\mathcal{A}(p)$.
\end{itemize}
\label{def:ae_pn}
\end{definition}

In this work, we refer to the types and placement of tokens in the network (plus the network tag, the clock, and the total reward) with the term \textit{marking}.

\begin{figure}[h!]
    \centering
    \includegraphics[width=\textwidth]{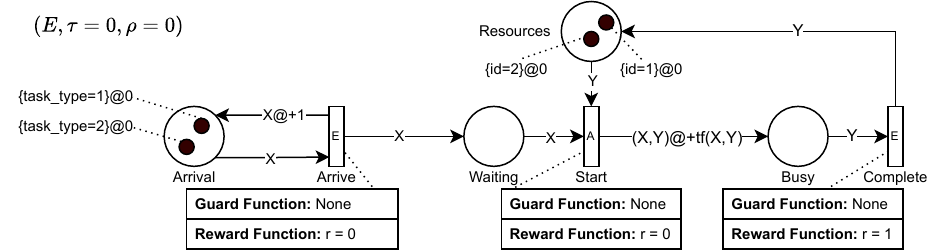}
    \caption{A task assignment problem expressed in A-E PN notation.}
    \label{fig:aepn_example}
\end{figure}

As an example, let us consider the A-E PN in~\cref{fig:aepn_example}, representing a task assignment problem with cases composed of a single activity and two available resources.

In~\cref{fig:aepn_example}, two tokens are used to represent cases that enter the system (two for each timestep). Two tokens represent resources that need to be assigned to the cases, and different assignments take different times to complete according to the function \(tf(x,y)\), defined as:

$$ tf(X, Y) =
\begin{cases}
1 & \text{if } X.\text{task\_type} = Y.\text{id} \\
2 & \text{otherwise}
\end{cases} $$

A reward of \(1\) is provided every time a case is completed. Thus, the objective is to minimize the cycle time of cases (to maximize the reward). 

\begin{figure}[h!]
    \centering
    \includegraphics[width=0.8\textwidth]{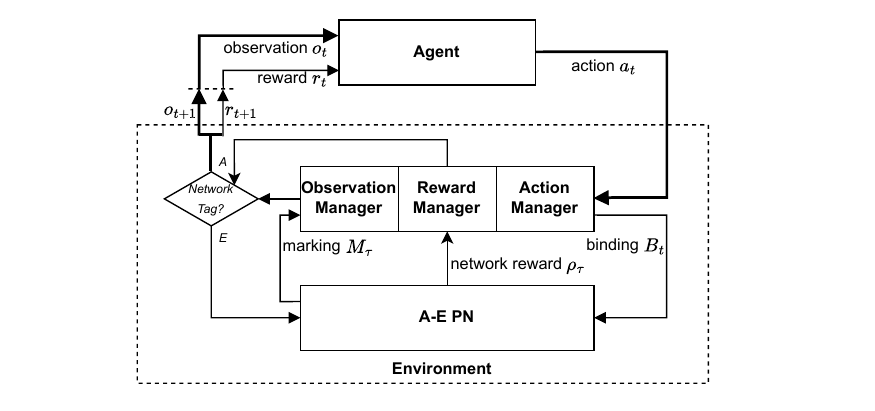}
    \caption{The classic reinforcement learning cycle integrated with A-E PN.}
    \label{fig:drl_cycle}
\end{figure}

The A-E PN relies on the periodic iteration between non-deterministic events, represented by \(E\) transitions and triggered when the network tag is \(E\), and sequential decision making, represented by the choice of which tokens' combinations to use to fire \(A\) transitions when the network tag is \(A\). The network tag and clock are updated as soon as there are no transitions of the current network tag that are enabled, and the system is in a deadlock (i.e., the simulation terminates) when no transitions are enabled in any of the two tags.
In~\cref{fig:aepn_example}, the network tag (first element of the tuple in the upper left corner), is \textit{E}, so only the \textit{Arrive} transition is enabled. After the arrival of both cases, the network tag will be updated to \textit{A}, enabling transition \textit{Start}.
In the context of A-E PN, a policy is a function that, given the PN with the relative \textit{marking} (i.e., the tokens contained in every place of the network at a given time), selects an action transition to \textit{fire} (i.e., activate), and the set of tokens to use to fire that transition (in Petri Net terms, a \textit{binding}). Such a function can be rule-based (what we call a heuristic), or it can be learned by a DRL agent. In~\cref{fig:drl_cycle} we report a visual representation of the interactions between the agent and the A-E PN, enriching the classic RL cycle~\cite{Sutton1998}.

In the figure, the A-E PN interacts with the agent every time an action needs to be taken. To this end, the current \textit{marking} of the A-E PN, representing the state of the system, is translated into a suitable \textit{observation} for the agent by encoding only the elements of the simulation that should be exposed to the decision-making algorithm, and the total \textit{reward} cumulated by previous actions is used as a measure of the goodness of the current policy. When provided with a new observation, the agent returns an \textit{action} (integer-valued) that corresponds to a specific binding on an action transition in the A-E PN.
The observation is an element of particular relevance to the framework and is defined in~\cite{marrella_universal_2024} as reported in~\cref{def:assignment_graph}.

\begin{definition}{\textbf{(Assignment Graph)}}
An assignment graph $G$ is a tuple $G = (V, D, Y, A, \phi, \theta)$ where: 
\begin{itemize}
 \item $V$ is a finite set of nodes.
 \item $D \subseteq V \times V$ is a set of ordered pairs of vertices, known as (directed) edges.
 \item $Y = \{ \text{A\_Transition}, \text{E\_Transition}, \dots \}$ is a finite set of node types, where $\text{A\_Transition}$ represents the action type and $\text{E\_Transition}$ represents the evolution type. Other node types represent the different places in the PN, discriminated on the base of their attributes. 
 \item $A$ is a finite set of attributes.
 \item $\phi: V \rightarrow Y$ is a function assigning a type to each node.
 \item $\theta: Y \rightarrow 2^A$ is a function assigning a set of attributes to each type.
\end{itemize} 
\label{def:assignment_graph} 
\end{definition}

Such a graph representation of the observations translates the action into the selection of a node on the graph (between the nodes corresponding to the bindings enabling the action transitions). The previous versions of the A-E PN framework assumed that all the elements of the state would be included in the assignment graph and that a single action transition would be present in the network. In the next section, we introduce the necessary elements to relax these assumptions.

\section{Method}\label{sec:method}

This section is dedicated to the two main novel elements introduced in GymPN. In particular, we dedicate one subsection to introducing the concept of observability and the GymPN approach to model it, and one to the GymPN modeling approach for environments with multiple action types.

\subsection{Process observability}
GymPN is a framework that combines elements of simulation with elements of automatic decision-making. When it is possible to expose all the elements of the state of the simulation to the agent, we speak of \textit{fully observable process}. Previous A-E PN implementations only considered this process class. In GymPN, we introduce a modeling mechanism that allows us to represent \textit{partially observable processes} by allowing single token attributes or entire places to be excluded from the observations presented to the agent.

We can formally introduce the concept of observability in a revised version of~\cref{def:ae_pn} by adding two more elements to the tuple representing A-E PN.

\begin{sloppypar}
\begin{definition}[Partially Observable Action-Evolution Petri Net] A Partially Observable Action-Evolution Petri Net (A-E PN) is a tuple $AEPN =(A, P, T, F, \mathcal{A}, G, E, I, L, l_0, \mathcal{R}, \rho_0, \mathcal{O}_{P}, \mathcal{O}_{\mathcal{A}})$, where $(A, P, T, F, \mathcal{A}, G, E, I, L, l_0, \mathcal{R}, \rho_0)$ follow~\cref{def:ae_pn}, and:
\begin{itemize}
\item \(\mathcal{O}_{P}: P \rightarrow \{0, 1\}\) is a \textit{place observability function} that maps every place in \(P\) to a boolean value, where \(0\) indicates that the place is observable and \(1\) indicates that the place is unobservable.
\item \(\mathcal{O}_{\mathcal{A}} : \{ (p, a) \mid p \in P,\, a \in \mathcal{A}(p) \} \rightarrow \{0, 1\}
\) is an \textit{attribute observability function} that maps, for each place \(p\), the attributes \(a \in \mathcal{A}(p)\) to a boolean value, where \(0\) indicates that the attribute is observable and \(1\) indicates that the attribute is unobservable.
\end{itemize}
\label{def:po_ae_pn}
\end{definition}
\end{sloppypar}

The adjustments introduced in~\cref{def:po_ae_pn} are enough to enable the A-E PN framework to express partially observable processes. The newly defined partially observable A-E PN retains the expressive power to represent the simulated environment and the decision-making problem in a single notation, while allowing for the separation of the elements that pertain to the former from the ones that pertain to the latter. As an example, we can consider the problem presented in~\cref{fig:aepn_example}. In such a problem, it would be reasonable to exclude the arrival process from the elements that are included in the observation. Given the newly introduced elements in~\cref{def:po_ae_pn}, this can be achieved simply by setting \(\mathcal{O}_{P}\) so that \(\mathcal{O}_{P}(\textit{Arrival}) = 1\). In the next subsection, we provide the details of the algorithm that achieves this effect.

\subsection{Multiple actions}
In previous implementations of the A-E PN framework, the focus was on modeling problems with a single action transition. However, in many cases, complex processes require the making of decisions at different stages of the process. In this section, we show how this limitation is lifted in GymPN with minimal modifications to the existing theory of A-E PN.

\begin{figure}[h!]
    \centering
    \includegraphics[width=0.95\textwidth]{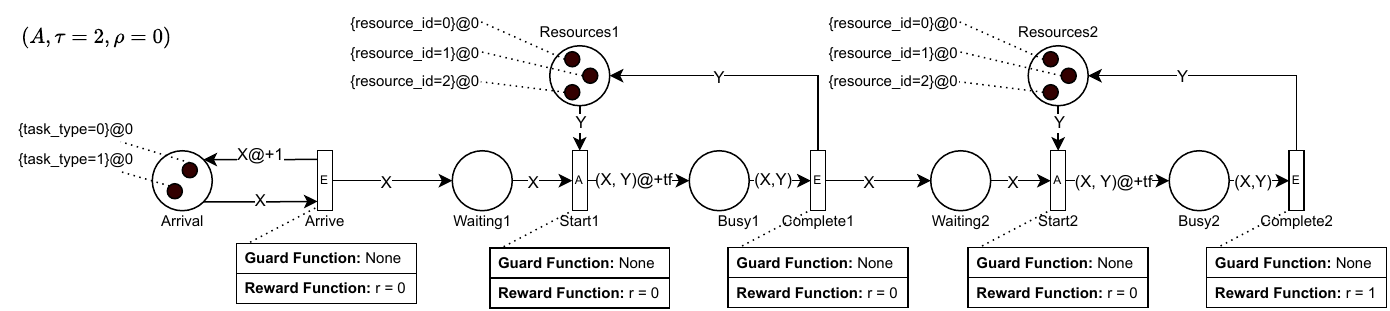}
    \caption{An example of A-E PN with disjoint actions.}
    \label{fig:disjoint_actions}
\end{figure}

\begin{figure}[h!]
    \centering
    \includegraphics[width=\textwidth]{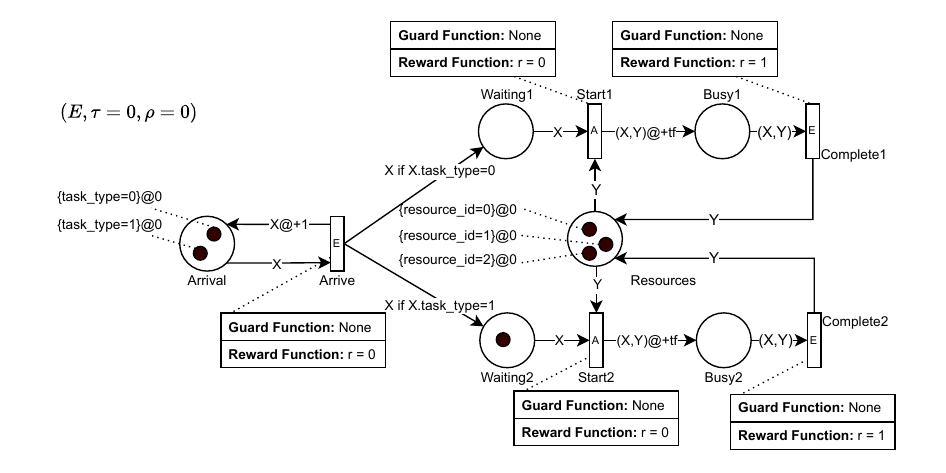}
    \caption{An example of A-E PN with joint actions.}
    \label{fig:joint_actions}
\end{figure}

To introduce the problem of modeling multiple actions, we refer to the two business process decision-making problem patterns in~\cref{fig:disjoint_actions} and~\cref{fig:joint_actions}. The two images present two archetypal cases of systems with multiple actions. The problem in~\cref{fig:disjoint_actions} includes two action transitions (\textit{Start1} and \textit{Start2}) that do not share any incoming place. We refer to this problem as having \textit{disjoint actions}. On the other hand,~\cref{fig:joint_actions} presents two action transitions (\textit{Start1} and \textit{Start2}) that share a common incoming place (\textit{Resources}). We refer to this problem as having \textit{joint actions}.

\begin{definition}[Mapping from Petri Net to GymPN Assignment Graph]
A mapping from (expanded) A-E PN to assignment graph is a function from a marked A-E PN $\mathcal{T}$ to an assignment graph $G$ having one node for each place or transition in $\mathcal{T}$ and one (directed) edge for each arc in $\mathcal{T}$ to the corresponding source and destination nodes in $G$.
\end{definition}

At a high level, the mapping function operates in three main steps:
\begin{enumerate}
    \item \textbf{Place mapping}: every non-empty place is mapped to a node with the same attributes as the single token in the original place, plus its time.
    \item \textbf{Transition mapping}: every transition is mapped to a node without attributes.
    \item \textbf{Arc mapping}: every arc not connected to empty places is mapped to an edge connecting nodes corresponding to the arc's source and destination.
\end{enumerate}

The mapping algorithm from A-E PN \(\mathcal{T}\) to assignment graph \(G\) is described in~\cref{alg:translation_gympn}. In the algorithm, we use the notation $v(t)$ to refer to the value of attributes of token $t$, $\tau (t)$ to refer to its time, $v(n')$ to refer to the values associated with attributes of the type of the newly created node $n'$, and $p(t)$ to refer to a place $p$ marked with a single token $t$.  We also assume the existence of the functions \textit{get\_A\_transition\_type} and \textit{get\_E\_transition\_type}, which map the action and evolution transitions in the expanded network to an integer value indicating the original transition in the non-expanded network. A third function \textit{one\_hot} is used to map the integer value to its one-hot encoding value. Finally, we introduce two support multi-sets, $H_p$ and $H_t$, used to store tuples of nodes and transitions in the A-E PN and the corresponding edges in the assignment graph.

\begin{algorithm}
\caption{Petri Net to GymPN Assignment Graph Mapping}
\label{alg:translation_gympn}
\begin{algorithmic}[1]
\Procedure{map}{$\mathcal{T}$}
\State $V_G \gets \emptyset$, $D_G \gets \emptyset$, $Y_G \gets \emptyset$, $A_G \gets A_T$, $H_p \gets \emptyset$, $H_t \gets \emptyset$
\For{$\forall p \in P_{\mathcal{T}}$}
    \If{$\mathcal{O}_P(p) = 0 \land \exists t \text{ s.t. } p(t)$}
        \State $\mathcal{A}_{obs}(p) \gets \{a \in \mathcal{A}(p) \mid \mathcal{O}_{\mathcal{A}}(p, a) = 0\}$
        \State $v_{obs}(t) \gets \{v(t)[a] \mid a \in \mathcal{A}_{obs}(p)\}$
        \State $V_G \gets V_G \cup \{n' \text{ s.t. } \phi(n') = \mathcal{A}_{obs}(p) \land v(n') = \{\tau (t)\} \cup v_{obs}(t) \}$
        \State $H_p \gets H_p \cup (p, n')$
    \EndIf
\EndFor
\For{$\forall tr \in T_{\mathcal{T}}$}
    \If{$L_{\mathcal{T}}(tr) = \text{'E'}$}
        \State $type \gets \text{get\_E\_transition\_type}(tr)$
        \State $v_{onehot} \gets \text{one\_hot}(type, \text{num\_E\_types})$
        \State $V_G \gets V_G \cup \{n' \text{ s.t. } \phi(n') = \text{E\_Transition} \land v(n') = v_{onehot}\}$
    \Else
        \State $type \gets \text{get\_A\_transition\_type}(tr)$
        \State $v_{onehot} \gets \text{one\_hot}(type, \text{num\_A\_types})$
        \State $V_G \gets V_G \cup \{n' \text{ s.t. } \phi(n') = \text{A\_Transition} \land v(n') = v_{onehot}\}$
    \EndIf
    \State $H_{t} \gets H_{t} \cup (tr, n')$
\EndFor
\For{$\forall a \in F_T$}
    \If{$a[0] \in P_{\mathcal{T}} \land \mathcal{O}_P(a[0]) = 0 \land \mathcal{O}_P(a[1]) = 0$}
        \State $D_G \gets D_G \cup \{(H_p[a[0]], H_{t}[a[1]])\}$
    \ElsIf{$a[1] \in P_{\mathcal{T}} \land \mathcal{O}_P(a[1]) = 0 \land \mathcal{O}_P(a[0]) = 0$}
        \State $D_G \gets D_G \cup \{(H_{t}[a[0]], H_p[a[1]])\}$
    \EndIf
\EndFor
\State \textbf{return} $G$
\EndProcedure
\end{algorithmic}
\end{algorithm}

The new mapping in~\cref{alg:translation_gympn} excludes all unobservable places and attributes from the assignment graph. Moreover, it ensures that the information regarding the original transitions is available in one-hot encoded form, thus enabling the solving algorithm to discriminate between different action transitions.

\begin{figure}[h!]
    \centering
    \includegraphics[width=\textwidth]{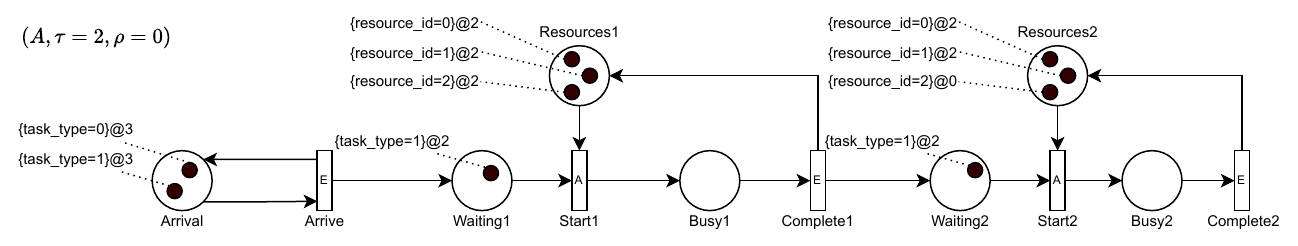}
    \caption{A-E PN with disjoint actions.}
    \label{fig:disjoint_initial}
\end{figure}

To better clarify the procedure in~\cref{alg:translation_gympn}, let us consider the two problems presented in~\cref{fig:disjoint_actions} and~\cref{fig:joint_actions}. Since handling unobservable features and tokens is relatively straightforward, we focus on use cases with multiple actions.
In the case of disjoint actions (~\cref{fig:disjoint_actions}), we can consider the marking reported in~\cref{fig:disjoint_initial} (simplified for improved readability). Two activities are ready to be assigned (places \textit{Wait1} and \textit{Wait2}). All resources are available.

\begin{figure}[h!]
    \centering
    \includegraphics[width=\textwidth]{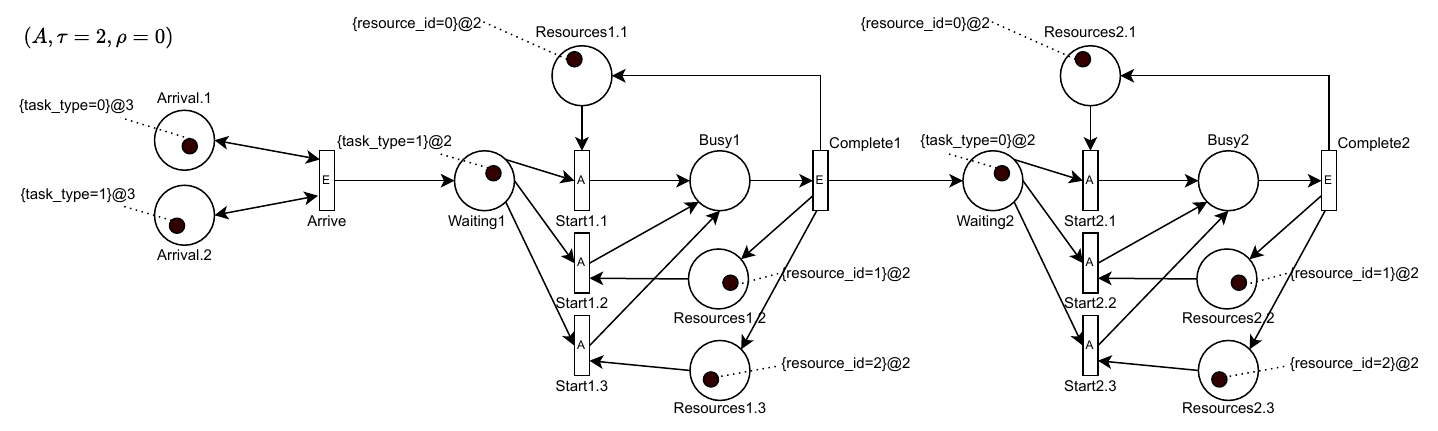}
    \caption{Expanded A-E PN with disjoint actions.}
    \label{fig:disjoint_expanded}
\end{figure}

The expansion procedure would produce the A-E PN reported in~\cref{fig:disjoint_expanded}, where every place contains a single token.

\begin{figure}[h!]
    \centering
    \includegraphics[width=\textwidth]{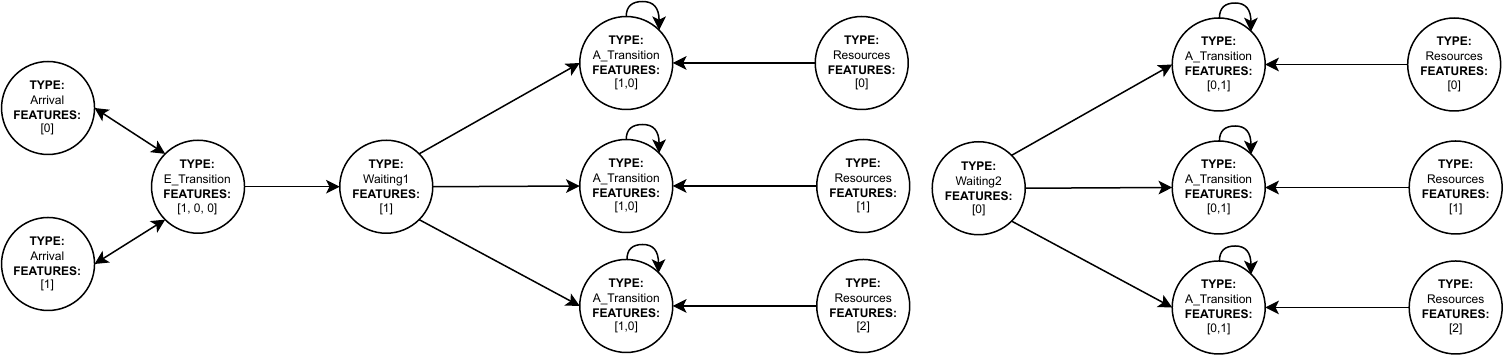}
    \caption{Assignment graph observation for A-E PN with disjoint actions.}
    \label{fig:disjoint_graph}
\end{figure}

The GymPN mapping algorithm applied to the A-E PN in~\cref{fig:disjoint_expanded}, produces the assignment graph in~\cref{fig:disjoint_graph}.

\begin{figure}[h!]
    \centering
    \includegraphics[width=\textwidth]{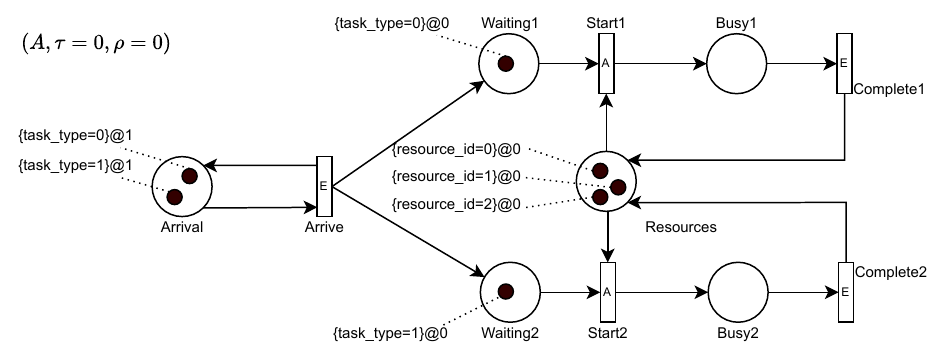}
    \caption{A-E PN with joint actions.}
    \label{fig:joint_initial}
\end{figure}

In the same way, we can derive the assignment graph for an A-E PN with joint actions (~\cref{fig:joint_actions}). To this end, let us consider the marking reported in~\cref{fig:joint_initial}.

In this case, the expansion procedure would produce the A-E PN reported in~\cref{fig:disjoint_expanded}.

\begin{figure}[h!]
    \centering
    \includegraphics[width=0.95\textwidth]{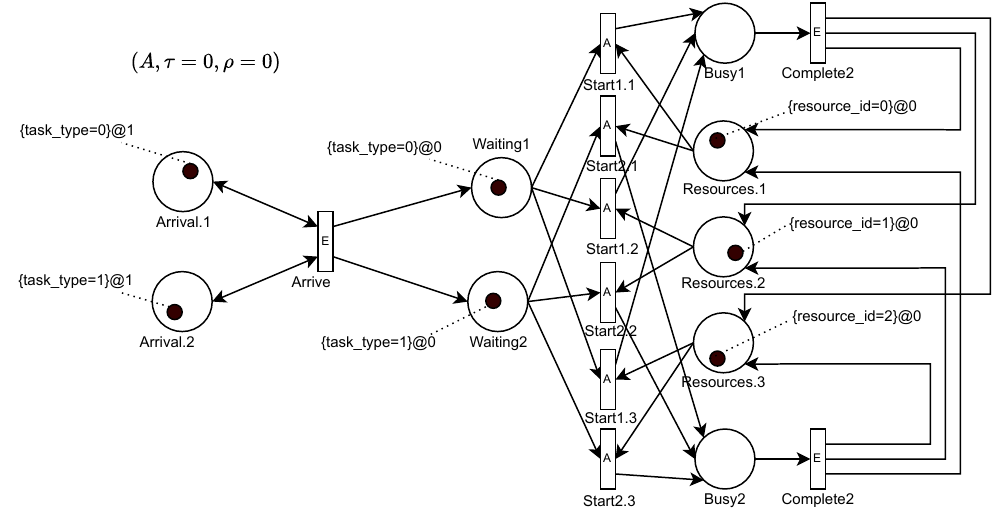}
    \caption{Expanded A-E PN with joint actions.}
    \label{fig:joint_expanded}
\end{figure}

The GymPN mapping algorithm applied to the A-E PN in~\cref{fig:disjoint_expanded}, produces the assignment graph in~\cref{fig:disjoint_graph}.

\begin{figure}[h!]
    \centering
    \includegraphics[width=0.8\textwidth]{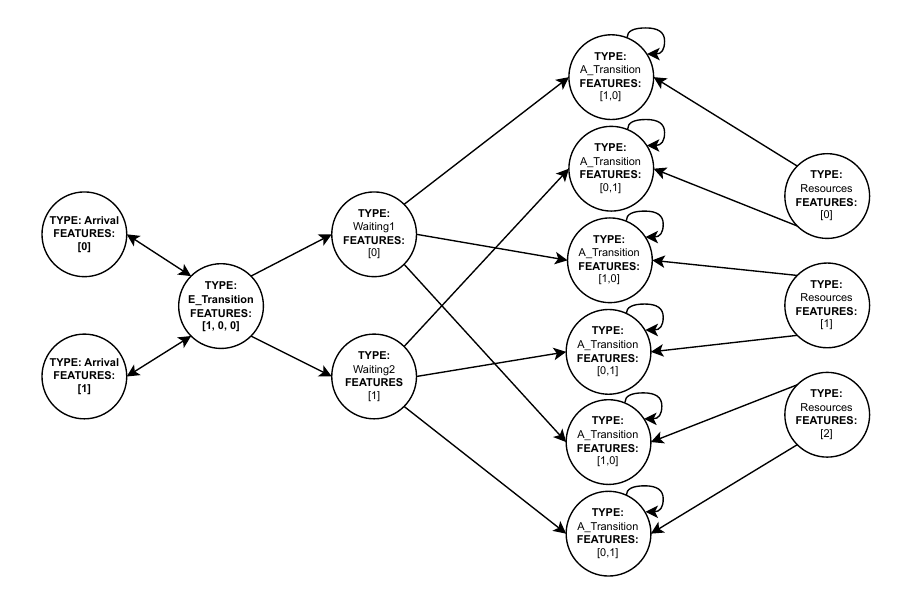}
    \caption{Assignment graph observation for A-E PN with joint actions.}
    \label{fig:joint_graph}
\end{figure}

The assignment graphs in the two considered cases clearly discriminate different action types, allowing the policy network to distinguish different action nodes. It can be noticed that, in~\cref{fig:disjoint_graph}, two disconnected graphs are produced. However, this is not a problem: due to the nature of the Graph Neural Networks used to approximate the policy and value functions, the two graphs will be considered as a single observation, and only the encoding of action nodes will be used to determine the chosen action. The mechanism allowing this desirable property is described in detail in~\cite{marrella_universal_2024}. 

\section{Implementation}\label{sec:implementation}

\begin{figure}[h!]
    \centering
    \includegraphics[width=0.9\textwidth]{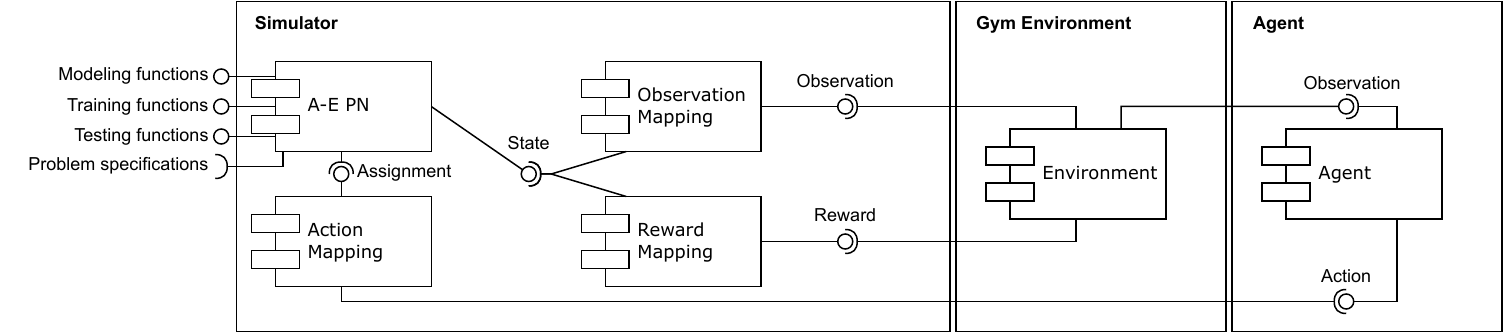}
    \caption{The GymPN component model.}
    \label{fig:component_model}
\end{figure}

GymPN is a Python library that implements the partially observable A-E PN definition, as well as a collection of utilities to facilitate modeling the desired problem, as well as training and testing DRL and heuristic policies. It consists of three main modules, as illustrated in~\cref{fig:component_model}: the \emph{Simulator} that can be used to model and simulate a business process decision-making problem, the \emph{Gym Environment} that encapsulates the simulator as a DRL problem, enabling it to be solved using the Gymnasium library, and the \emph{Agent} that learns the policy that solves the business process decision-making problem.

The Simulator facilitates modeling and simulating a business process decision-making problem. To this end, the simulator contains four specialized modules: A-E PN, Observation Mapping, Reward Mapping, and Action Mapping.

The A-E PN module implements the A-E PN framework described throughout this work, exposing to the user the primitives that are necessary to define the desired business process decision-making problem, and to train and test DRL or heuristic policies. In order to define such a problem, the user is expected to define the corresponding A-E PN in terms of its places, transitions (with relative guard, behavior, and reward functions), and the marking (by adding the desired tokens to the places). Optionally, unobservable places and token attributes can be defined in a similar fashion.

The observation mapping module is responsible for translating the A-E PN state (i.e., its marking) into the assignment graph observation, excluding unobservable elements. 
The action mapping module is responsible for translating the integer-valued actions returned by the agent into suitable bindings to fire in the A-E PN.
The reward mapping module keeps track of the total reward accumulated during a trajectory and returns the correct reward every time an action transition fires. 

GymPN implements a Gym environment as a communication layer between the simulator and the agent. The environment module ingests the observations and rewards from the simulator. The former is passed to the agent to perform actions, while the second is used to store information regarding the goodness of the current policy, allowing it to improve during the training.

The agent module represents the decision-making algorithm that encodes the policy. Given an observation, it produces a single integer-valued action. Currently, GymPN implements the well-known Proximal Policy Optimization (PPO) algorithm, while other DRL algorithms will be introduced later.

The GymPN library is available at \url{https://github.com/bpogroup/gympn}.

\section{Evaluation}\label{sec:evaluation}
The evaluation of the proposed approach relies on business process decision-making problem patterns that demonstrate the possibility of learning optimal policies when multiple actions are present in the problem.

In this section, we consider the problem of task assignment, involving the decision of which resource to allocate to which activity under which circumstances (i.e., in which state of the process). To this end, we consider a set of eight business process decision-making problem patterns. The considered patterns cover the most important workflow patterns reported in ~\cite{van_der_aalst_workflow_2003}, namely:
\begin{enumerate}
    \item \textbf{Sequence}: activities in a case appear sequentially.
    \item \textbf{Parallelism}: two or more activities in a case need to be processed in parallel for the case to continue.
    \item \textbf{Arbitrary Cycle}: a subset of activities can be repeated multiple times in a case.
    \item \textbf{Exclusive Choice}: two or more activities in a case appear as mutually exclusive.
\end{enumerate}

\begin{figure}[h!]
    \centering
    \includegraphics[width=\textwidth]{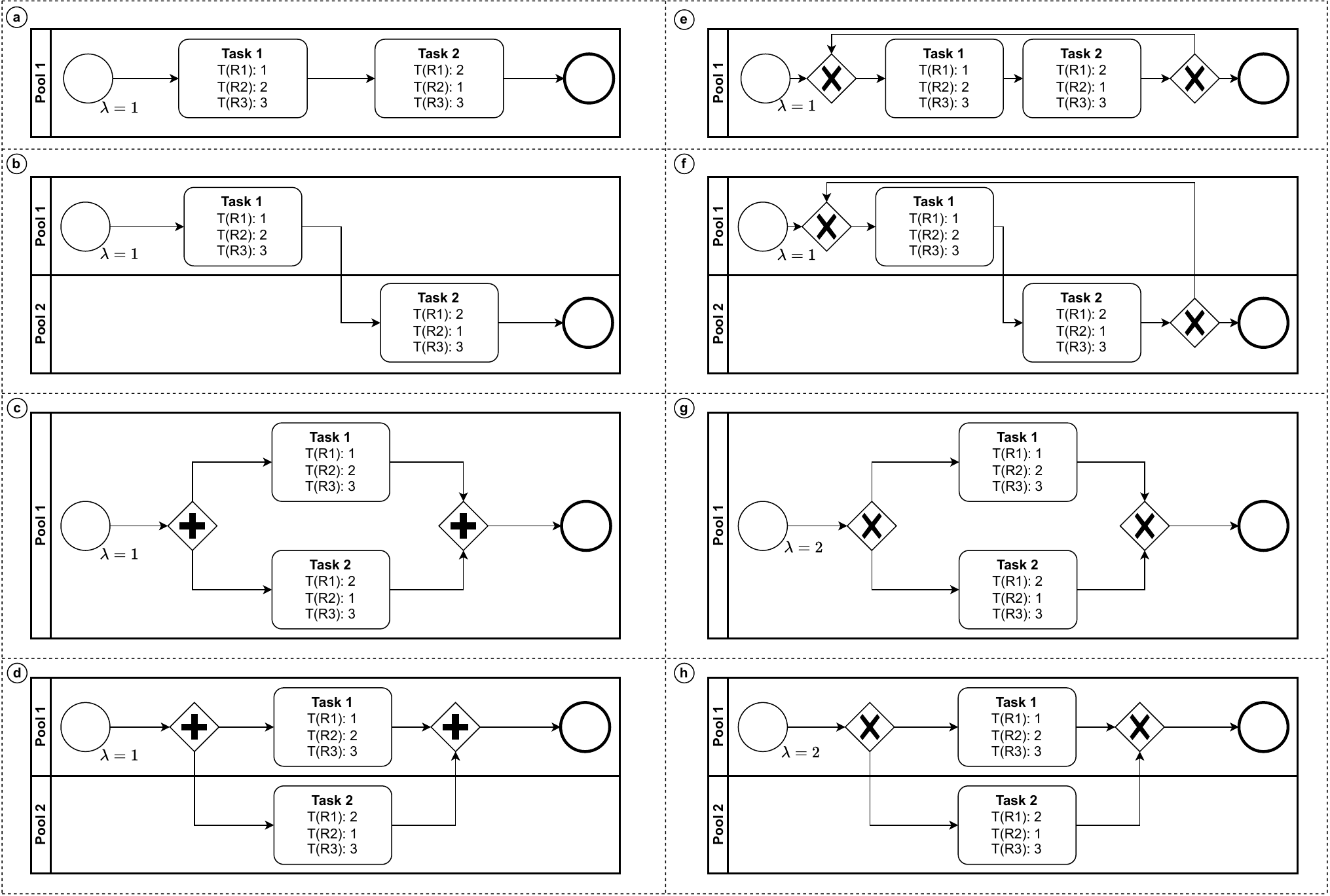}
    \caption{BPMN diagrams of the eight example problems used for evaluation.}
    \label{fig:evaluation_bpmn}
\end{figure}

To validate the novel mapping algorithm introduced in~\cref{alg:translation_gympn}, we propose two variations for each of the cited patterns: one with joint actions (i.e., with a single resource pool), and one with disjoint actions (i.e., with separate resource pools). 
In~\cref{fig:evaluation_bpmn} we represent the eight example problems using BPMN. In the figure, problems \textit{(a)} and \textit{(b)} represent, respectively, sequences of joint and disjoint actions. Problems \textit{(c)} and \textit{(d)} represent, respectively, parallel joint and disjoint actions. Problems \textit{(e)} and \textit{(f)} represent, respectively, cycles of joint and disjoint actions. Problems \textit{(g)} and \textit{(h)} represent, respectively, exclusive joint and disjoint actions. 
For example, \cref{fig:evaluation_bpmn}.a shows a process with two activities that are executed in a sequence, where new cases arrive every time unit (i.e. with a rate $\lambda=1$). There is one pool with three resources $R_1, R_2, R_3$ that can perform both activities, but perform differently on each actitity. For example, resource $R_1$ takes 1 time unit to perform activity 1, but 2 time units to perform activity 2. The objective is to maximize the number of cases processed in a trajectory of \(10\) time units (or, equivalently, to minimize the cycle time of cases).

In problems \textit{(a)}, \textit{(b)}, \textit{(e)}, \textit{(f)}, \textit{(g)}, and \(h\) cases are composed of a single activity which needs to be executed multiple times, while in problems \textit{(c)} and \textit{(d)} cases are composed of two activities, one of each type. The resource pools contain three resources, each characterized solely by a \textit{resource\_id} attribute, similar to~\cref{fig:disjoint_actions} and~\cref{fig:joint_actions}. An arrival rate of \(1\) is assumed for every problem, except for problems \textit{(g)} and \textit{(d)} where the arrival rate is raised to \(2\) to balance higher resource availability. Each resource has a different completion time for each task type (reported in the activities in~\cref{fig:evaluation_bpmn}). It is easy to see that, in every problem, the algorithm is expected to always assign resource \(1\) to activities of type \(1\) and resource \(2\) to activities of type \(2\), while resource \(3\) should remain unused because it takes \(3\) time units to complete any activity.

In~\cref{tab:results} we report the mean and standard deviation of the cumulative reward over \(10\) episodes of \(10\) time units following a random policy and following the (deterministic) trained policy.

\begin{table}[ht]
\centering
\small
\begin{tabular}{>{\centering\arraybackslash}m{1.5cm} >{\centering\arraybackslash}m{2.5cm} >{\centering\arraybackslash}m{2.5cm} >{\centering\arraybackslash}m{2.5cm}}
\toprule
\textbf{Problem} & \textbf{Random} & \textbf{PPO} & \textbf{Optimum} \\
\midrule
(a) & 6.1 \( \pm \) 0.8 & 9 \( \pm \) 0 & \textbf{9} \\
(b) & 7.5 \( \pm \) 0.5 & 9 \( \pm \) 0 & \textbf{9} \\
(c) & 6.8 \( \pm \) 0.4 & 10 \( \pm \) 0 & \textbf{10} \\
(d) & 8.8 \( \pm \) 0.4 & 10 \( \pm \) 0 & \textbf{10} \\
(e) & 2.9 \( \pm \) 1.4 & 9 \( \pm \) 0 & \textbf{9} \\
(f) & 3.9 \( \pm \) 0.4 & 9 \( \pm \) 0 & \textbf{9} \\
(g) & 15.1 \( \pm \) 1.1 & 20 \( \pm \) 0 & \textbf{20} \\
(h) & 17.8 \( \pm \) 1.2 & 20 \( \pm \) 0 & \textbf{20} \\
\bottomrule
\end{tabular}
\caption{Performance comparison between Random, PPO, and the Optimum}
\label{tab:results}
\end{table}

In all instances, the DRL agent learned the optimal policy. Using the deterministic policy (that is, taking action with higher probability instead of sampling from the probability distribution that is the output of the policy network) in problems that do not include stochastic elements results in a variance of \(0\).

\section{Conclusion}\label{sec:conclusion}

This work introduced GymPN, a library for automated decision-making in process management systems. GymPN is built on top of SimPN, a library for simulation with Petri Nets, and the two share the same syntax and basic constructs. Compared to previous versions of the A-E PN framework, GymPN introduces the possibility of expressing partially unobservable processes and handling multiple action transitions. The first novelty required an addition to the attributed A-E PN framework, while the latter required a modification of the expansion algorithm. The novel elements were evaluated by means of a set of problem instances representing the basic workflow patterns in business processes. In all instances, a DRL agent was capable of learning the optimal assignment policy.
At the time of writing, the GymPN library still supports only single-agent DRL, and only the PPO algorithm is available as part of the code. In the future, support for multi-agent DRL will be provided, as well as other DRL algorithms. The reason to include different algorithms is motivated by the relatively low data-efficiency of PPO, while alternatives like rollout-based algorithms proved to be more efficient, especially in highly stochastic environments. While such features will be implemented directly by the authors, we remain confident that the research community will provide further suggestions for improvements. 
\bibliographystyle{splncs04}
\bibliography{springer_formatted}

\end{document}